\documentclass{amsart}
\usepackage{mathtools}
\usepackage{amssymb}
\usepackage{hyperref}
\usepackage{biblatex}
\usepackage{algorithm}
\usepackage{algorithmic}
\usepackage{xifthen} 

\hypersetup{
    colorlinks,
    linkcolor={red},
    citecolor={green},
    urlcolor={blue}
}

\makeatletter
\@namedef{subjclassname@2020}{\textup{2020} Mathematics Subject Classification}
\makeatother

\theoremstyle{definition}
\newtheorem{defn}{Definition}
\newtheorem{example}{Example}

\providecommand{\set}[2][]{
	\ifthenelse{\isempty{#1}}{
		\left\{#2\right\}
	}{
		\left\{\,#1\;\middle|\;#2\,\right\}}
	}
\DeclareMathOperator{\pow}{Sb} 
\DeclareMathOperator{\id}{Id} 
\DeclareMathOperator{\op}{Op} 
\DeclareMathOperator{\proj}{Proj} 
\providecommand{\N}{\mathbb{N}} 
\providecommand{\PP}{\mathbb{P}} 
\providecommand{\Z}{\mathbb{Z}} 
\providecommand{\R}{\mathbb{R}} 
\providecommand{\F}{\mathbb{F}} 
\DeclarePairedDelimiter\abs{\lvert}{\rvert} 
\newcommand{\norm}[1]{\left\lVert#1\right\rVert} 
\DeclareMathOperator{\enm}{End} 
\DeclareMathOperator{\mat}{Mat} 
\DeclareMathOperator{\ham}{\mathbf{Ham}} 
\DeclareMathOperator{\poly}{Poly} 
\DeclareMathOperator{\im}{Im} 
\DeclareMathOperator{\dihedralend}{Dihed} 
\DeclareMathOperator{\swapend}{Swap} 
\DeclareMathOperator{\blankend}{Blank} 
\DeclareMathOperator{\multind}{MultInd} 
\DeclareMathOperator{\minc}{\mathbf{MinC}} 

\begin{document}
\title[Discrete neural nets and polymorphic learning]{Discrete neural nets and polymorphic learning}
\author[C. Aten]{Charlotte Aten}
\address{Department of Mathematics\\
University of Denver\\Denver 80208\\USA}
\urladdr{\href{https://aten.cool}{https://aten.cool}}
\email{\href{mailto:charlotte.aten@du.edu}{charlotte.aten@du.edu}}
\thanks{This project was supported by NSF Grant US NSF HDR TRIPODS 1934962. Thanks to Nick Cimaszewski and Andrey Yao for their participation in the corresponding REU program. Thanks to Rachel Dennis for the code she contributed as part of her senior thesis project\cite{dennis2023}.}
\subjclass[2020]{68T07, 08A70, 05C60}
\keywords{Neural nets, polymorphisms, clones, universal algebra, machine learning}

\begin{abstract}
Theorems from universal algebra such as that of Murskiĭ from the 1970s have a striking similarity to universal approximation results for neural nets along the lines of Cybenko's from the 1980s. We consider here a discrete analogue of the classical notion of a neural net which places these results in a unified setting. We introduce a learning algorithm based on polymorphisms of relational structures and show how to use it for a classical learning task.
\end{abstract}

\maketitle

\tableofcontents

\section{Introduction}
This work has its genesis in the observation that a class of theorems from universal algebra (exemplified by Murskiĭ's Theorem from the 1970s\cite[Section 6.2]{bergman2012}) were discrete forerunners of a class of theorems from machine learning (exemplified by Cybenko's work from the 1980s\cite{cybenko1989}). The similarity between these two types of results motivates the mathematical analysis of a discrete analogue of the usual (continuous) treatment of neural nets. Unsurprisingly, this analogue fits neatly into the established theory of universal algebra, and in particular the theory of clones. Although our work has no immediate predecessor to our knowledge, we direct the reader to a similar perspective in evolutionary computation\cite{clark2013} as well as the successful use of polymorphisms of relational structures in the resolution of the Dichotomy Conjecture for the Constraint Satisfaction Problem by Bulatov in 2017\cite{bulatov2017}. See \cite{jeavons1998} for an introduction to such applications of universal algebra and clones to combinatorial problems. The equivariant maps of equivariant neural nets\cite{lim2022} can be viewed as polymorphisms, although existing work on such neural nets does not appear to have moved in the direction we describe here.

We use Shalev-Shwartz and Ben-David as a general reference for the mathematical treatment of machine learning\cite{shalev-shwartz2014} and we use Bergman as a reference for universal algebra and clone theory\cite{bergman2012}.

The structure of the paper is as follows: We give some relevant background on universal algebra, clones, and polymorphisms, after which we introduce our concept of a discrete neural net. The reader need not be familiar with the established theory of neural nets nor their application in order to follow this section, although one who is will note that our concept of neural net both mathematically subsumes the established (continuous) one and also is, in a sense, closer to neural nets as they are actually implemented in digital computers. We then describe a learning algorithm which makes use of polymorphisms pertaining to the task at hand. Finally, we illustrate how this algorithm may be used for image classification and transformation.

In this paper we adopt the convention that the natural numbers are \(\N\coloneqq\set{0,1,2,\dots}\) and the positive integers are \(\PP\coloneqq\set{1,2,3,\dots}\). We write \(\pow(A)\) the indicate the power set of a set \(A\), we denote the cardinality of a set \(A\) by \(\abs{A}\), and we define
    \[
        \pow_\PP(A)\coloneqq\set[U\subset A]{\abs{U}\in\PP}.
    \]
Given \(n\in\PP\) we set \([n]\coloneqq\set{0,1,\dots,n-1}\). We denote by \(\mat_n(\F)\) the set of \(n\times n\) matrices over a field \(\F\). Our matrices are \(0\)-indexed, so the entries of a matrix \(a\in\mat_2(\F)\) are \(a_{00}\), \(a_{01}\), \(a_{10}\), and \(a_{11}\). This is to say that
    \[
        \mat_n(\F)=\F^{[n]^2}.
    \]
We indicate by \(\Sigma_n\) the set of permutations of \([n]\) and we indicate by \(\mathbf{\Sigma}_n\) the corresponding permutation group. We define \(\mathbf{D}_n\) to the the group of diheadral symmetries of an \(n\)-gon and write \(D_n\) to indicate the set of such symmetries. Note that \(\abs{D_n}=2n\).

\section{Universal algebra and clones}
We will make use of some of the language of universal algebra and the associated theory of clones. Algebras in the universal algebra sense are built from operations, which we will use as activation functions in our neural nets.

\begin{defn}[Operation, arity]
Given a set \(A\) and some \(n\in\N\), we refer to a function \(f\colon A^n\to A\) as an \emph{operation} on the set \(A\). We sometimes say that such a function is an \emph{\(n\)-ary operation} on \(A\) or that \(f\) has \emph{arity} \(n\).
\end{defn}

Although we won't make extensive use of the notion here, the basic objects of study for universal algebra are defined as follows.

\begin{defn}[Algebra, universe, signature, basic operation]
An algebra \(\mathbf{A}\coloneqq(A,F)\) consists of a set \(A\) (the \emph{universe} or \emph{underlying set} of \(\mathbf{A}\)) along with a collection \(F\coloneqq\{f_i\}_{i\in I}\) of operations on \(A\) indexed by a set \(I\). There exists a well-defined function \(\rho\colon I\to\N\) such that \(f_i\) has arity \(\rho(i)\). This function \(\rho\) is called the \emph{signature} of \(\mathbf{A}\). Each operation \(f_i\) is said to be a \emph{basic operation} of \(\mathbf{A}\).
\end{defn}

As usual in the categorical perspective on mathematics, one actually wants to study the morphisms between the objects under consideration. The relevant notion of morphism follows.

\begin{defn}[Homomorphism (of algebras)]
Given algebras \(\mathbf{A}\coloneqq(A,F)\) where \(F\coloneqq\{f_i\}_{i\in I}\) and \(\mathbf{B}\coloneqq(B,G)\) where \(G\coloneqq\{g_i\}_{i\in I}\), both of signature \(\rho\colon I\to\N\), we say that a function \(h\colon A\to B\) is a \emph{homomorphism} from \(\mathbf{A}\) to \(\mathbf{B}\) when for each \(i\in I\) we have for all \(a_1,\dots,a_{\rho(i)}\in A\) that
    \[
        h(f(a_1,\dots,a_{\rho(i)}))=g(h(a_1),\dots,h(a_{\rho(i)})).
    \]
When \(h\) is a homomorphism from \(\mathbf{A}\) to \(\mathbf{B}\) we write \(h\colon\mathbf{A}\to\mathbf{B}\).
\end{defn}

These homomorphisms will be generalized in \autoref{sec:polymorphism} and we will spend much of the rest of the paper discussing that generalization.

Given an algebra \(\mathbf{A}\coloneqq(A,F)\), one might notice that the basic operations could be composed in a way analogous to that of functions (of a single variable). This is captured in the next definition.

\begin{defn}[Generalized composite]
Given a set \(A\), a \(k\)-ary operation \(f\) on \(A\) and a collection \(\{g_i\}_{i\in\set{1,2,\dots,k}}\) of \(n\)-ary operations on \(A\) the \emph{generalized composite} of these operations is the \(n\)-ary operation
    \[
        f[g_1,\dots,g_k]\colon A^n\to A
    \]
where for any \(a\in A^n\) we set
    \[
        f[g_1,\dots,g_k](a)\coloneqq f(g_1(a),\dots,g_k(a)).
    \]
\end{defn}

Just as the notion of a monoid may arise from studying sets of functions closed under composition, so we arrive at the notion of a clone by studying sets of operations closed under generalized composition. Just as we would like the include the identity function in any monoid of functions under composition, we would also like the include the following operations in any clone.

\begin{defn}[Projection]
Given \(k,n\in\PP\) with \(k\le n\) we define the \emph{\((n,k)\)-projection} operation \(\pi_{n,k}\colon A^n\to A\) by
    \[
        \pi_{n,k}(a_1,\dots,a_n)\coloneqq a_k.
    \]
\end{defn}

We are now ready to define a clone.

\begin{defn}[Clone]
Given a set \(A\) and a set of operations \(\mathcal{C}\) on \(A\), we say that \(\mathcal{C}\) is a \emph{clone} when \(\mathcal{C}\) is closed under generalized composition and \(\mathcal{C}\) contains all the projection operations on \(A\).
\end{defn}

There are a couple natural examples of clones.

\begin{example}
Let \(\op_n(A)\coloneqq A^{A^n}\) be the set of all \(n\)-ary operations on the set \(A\). The largest clone on \(A\) is
    \[
        \op(A)\coloneqq\bigcup_{n\in\N}\op_n(A),
    \]
which merely consists of all possible operations on \(A\).
\end{example}

\begin{example}
The clone
    \[
        \proj(A)\coloneqq\set[\pi_{n,k}\colon A^n\to A]{k,n\in\PP\text{ and }k\le n}
    \]
of all projection operations is the smallest clone on the set \(A\).
\end{example}

Intermediate between these two examples are the clones of polymorphisms we will see in the proceeding section.

\section{Polymorphisms}
\label{sec:polymorphism}
We need a notion which simultaneously generalizes homomorphism and operation. That is, we would like functions which are simultaneously homomorphisms and operations in an appropriate way. If we choose such operations for our activation functions, we will find that we can ensure our neural net is at every step representing a function which obeys some pre-defined structure related to our learning task.

These ideas can be formulated in a more abstract categorical context, but for the purposes of this paper we will introduce polymorphisms only for the kinds of relational structures studied in model theory, as this will suffice for our examples.

Relational structures are like algebras in the previous section, but they can also carry the following type of basic object.

\begin{defn}[Relation, arity]
Given a set \(A\) and \(n\in\PP\) we say that \(\theta\subset A^n\) is an \emph{\(n\)-ary relation} on \(A\). We also say that \(\theta\) has \emph{arity} \(n\).
\end{defn}

The general model-theoretic concept of a structure is as follows.

\begin{defn}[Structure, universe, basic operations/relations]
A \emph{structure} \(\mathbf{A}\coloneqq(A,F,\Theta)\) conisists of a set \(A\) (the \emph{universe} or \emph{underlying set} of the structure) as well as indexed collections \(F\coloneqq\{f_i\}_{i\in I}\) and \(\Theta\coloneqq\{\theta_j\}_{j\in J}\) of operations on \(A\) (the \emph{basic operations} of \(\mathbf{A}\)) and of relations on \(A\) (the \emph{basic relations} of \(\mathbf{A}\)). We require that the index sets \(I\) and \(J\) be disjoint.
\end{defn}

Observe that such a structure consists of an algebra \((A,F)\) along with a collection of relations on a the underlying set \(A\). We have a notion of signature for structures.

\begin{defn}[Signature]
Given a structure \((A,F,\Theta)\) there exists a well-defined function \(\rho\colon I\cup J\to\N\) such that each basic operation \(f_i\) has arity \(\rho(i)\) and each basic relation \(g_j\) has arity \(\rho(j)\).
\end{defn}

There is again a notion of morphism for structures of a given signature.

\begin{defn}[Homomorphism (of structures)]
Given structures \(\mathbf{A}\coloneqq(A,F,\Theta)\) and \(\mathbf{B}\coloneqq(B,G,\Psi)\) where \(F\coloneqq\{f_i\}_{i\in I}\), \(G\coloneqq\{g_i\}_{i\in I}\), \(\Theta\coloneqq\{\theta_j\}_{j\in J}\), and \(\Psi\coloneqq\{\psi_j\}_{j\in J}\), both of the same signature \(\rho\colon I\cup J\to\N\), we say that a function \(h\colon A\to B\) is a \emph{homomorphism} from \(\mathbf{A}\) to \(\mathbf{B}\) when \(h\colon(A,F)\to(B,G)\) is a homomorphism of algebras and for each \(j\in J\) we have for all \(a_1,\dots,a_{\rho(j)}\in A\) that if
    \[
        (a_1,\dots,a_{\rho(j)})\in\theta_j
    \]
then
    \[
        (h(a_1),\dots,h(a_{\rho(j)}))\in\psi_j.
    \]
\end{defn}

Given a signature \(\rho\colon I\cup J\) we have a category whose objects are such structures with signature \(\rho\) are whose morphisms are the aforementioned homomorphisms. In this category has all products, so in particular we have that the \(n^{\text{th}}\) direct power of any structure \(\mathbf{A}\) exists for any \(n\in\N\).

We can now define one of the titular concepts of this paper.

\begin{defn}[Polymorphism]
Given a structure \(\mathbf{A}\) we say that a homomorphism \(f\colon\mathbf{A}^n\to\mathbf{A}\) is a \emph{polymorphism} of \(\mathbf{A}\).
\end{defn}

If \(\mathbf{A}\coloneqq(A,F,\Theta)\) and \(f\colon\mathbf{A}^n\to\mathbf{A}\) is a polymorphism of \(\mathbf{A}\) we have that \(f\colon A^n\to A\) is an \(n\)-ary operation in the manner previously discussed. The polymorphisms of a given structure form a clone.

\begin{defn}[Clone of polymorphisms]
Given a structure \(\mathbf{A}\) the \emph{clone of polymorphisms} of \(\mathbf{A}\) is
    \[
        \poly(\mathbf{A})\coloneqq\set[f\colon\mathbf{A}^n\to\mathbf{A}]{f\text{ is a polymorphism and }n\in\N}.
    \]
\end{defn}

Since the set of polymorphisms of a structure \(\mathbf{A}\) is closed under generalized composition, we will see that neural nets with such operations as their activation functions always model polymorphisms of \(\mathbf{A}\).

\section{Discrete neural nets}
In order to apply results and theory pertaining to finite algebras we will consider a discrete, finite analogue of neural networks. We begin with a description of neural nets which encompasses both the traditional, continuous variant and our new variant.

\begin{defn}[Neural net]
A \emph{neural net} \((V_1,\dots,V_r,E,\Phi,\le)\) with \(r\) layers on a set \(A\) consists of
	\begin{enumerate}
		\item a finite digraph \((V,E)\) (the \emph{architecture} of the neural net),
		\item for each \(v\in V\setminus V_1\) a function \(\Phi(v)\colon A^{\rho(v)}\to A\) (the \emph{activation function} of \(v\)),
		\item for each \(i\in[r]\) a total ordering \(\le_i\) of \(V_i\) (the \emph{vertex ordering} of layer \(i\))
	\end{enumerate}
where
	\begin{enumerate}
		\item \(V\coloneqq\bigcup_{i=1}^rV_i\),
        \item \(V_i\cap V_j=\varnothing\) when \(i\neq j\),
		\item the only edges in \(E\) are from vertices in \(V_i\) to vertices in \(V_{i+1}\) for \(i<r\),
		\item \(\rho(v)\) is the indegree of \(v\) in \((V,E)\), and
		\item if \(i\neq r\) then every vertex \(v\in V_i\) has nonzero outdegree.
	\end{enumerate}
\end{defn}

The set \(A\) in the preceding definition is also referred to as the \emph{universe} or the \emph{underlying set} of the neural net in question. The case where \(A=\R\) is that of a traditional neural net. Our treatment of activation functions is a bit different from that in the existing literature in the sense that we allow for nullary (constant) activation functions \(\Phi(v)\colon A^0\to A\) and do not assume that each activation function is of the form \(f(w\cdot x)\) where \(x=(x_1,\dots,x_{\rho(v)})\) is the tuple of arguments of \(\Phi(v)\) and \(w=(w_1,\dots,w_{\rho(v)})\) is a tuple of constant weights. Indeed, there is no relevant analogue of a dot product for an arbitrary set. Should one wish to separate out the weights in this presentation of a neural net the weights \(w_i\) may be taken to be the output values of nullary activation functions.

By the total ordering \(\le_i\) on each \(V_i\) (which is necessarily finite since we assume that \((V,E)\) is a finite digraph) we can list the vertices of \(V_i\) as
    \[
        v_{i,1}<v_{i,2}<\cdots<v_{i,\abs{V_i}}.
    \]
Similarly, we will use the notation \(\phi_{i,j}\coloneqq\Phi(v_{i,j})\). Naturally we will write \(v_{ij}\) and \(\phi_{ij}\) rather than \(v_{i,j}\) and \(\phi_{i,j}\) when there is no chance of confusion.

\begin{example}
\label{ex:neural_net}
Consider a neural net
    \[
        N\coloneqq(V_1,V_2,V_3,E,\Phi,\le)
    \]
with \(3\) layers on the set \(\F_5=\set{0,1,2,3,4}\) where
    \[
        V_1\coloneqq\set{v_{11},v_{12},v_{13}},
    \]
    \[
        V_2\coloneqq\set{v_{21},v_{22},v_{23},v_{24}},
    \]
    \[
        V_3\coloneqq\set{v_{31},v_{32}},
    \]
    \begin{align*}
        E\coloneqq\{ & (v_{11},v_{21}),(v_{11},v_{22}),(v_{12},v_{22}),(v_{13},v_{22}),(v_{13},v_{23}), \\
        & (v_{21},v_{31}),(v_{22},v_{31}),(v_{22},v_{32}),(v_{23},v_{32}),(v_{24},v_{32})\},
    \end{align*}
    \[
        \phi_{21}\colon\F_5\to\F_5
    \]
is given by
    \[
        \phi_{21}(x)\coloneqq-x,
    \]
    \[
        \phi_{22}\colon\F_5^3\to\F_5
    \]
is given by
    \[
        \phi_{22}(x,y,z)\coloneqq x-y+z,
    \]
    \[
        \phi_{23}\colon\F_5\to\F_5
    \]
is given by
    \[
        \phi_{23}(x)\coloneqq-x,
    \]
    \[
        \phi_{24}\colon\F_5^0\to\F_5
    \]
is given by
    \[
        \phi_{24}()\coloneqq3,
    \]
    \[
        \phi_{31}\colon\F_5^2\to\F_5
    \]
is given by
    \[
        \phi_{31}(x,y)\coloneqq xy^2,
    \]
    \[
        \phi_{32}\colon\F_5^3\to\F_5
    \]
is given by
    \[
        \phi_{32}(x,y,z)\coloneqq xy-z,
    \]
and \(\le\coloneqq(\le_1,\le_2,\le_3)\) is the triple of total orders determined by
    \[
        v_{11}<v_{12}<v_{13},
    \]
    \[
        v_{21}<v_{22}<v_{23}<v_{24},
    \]
and
    \[
        v_{31}<v_{32}.
    \]
The architecture \((V,E)\) of \(N\) is pictured in \autoref{fig:neural_net_architecture}.

    \begin{figure}
        \centering
        \includegraphics[height=5cm]{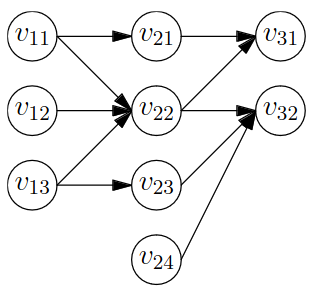}
        \caption{The architecture of a neural net}
        \label{fig:neural_net_architecture}
    \end{figure}
\end{example}

While all neural nets as we have defined them must have a finite number of nodes and edges in their architectures, the preceding example is of a neural net on a finite set, which is the type of neural net we will primarily examine here.

\begin{defn}[Discrete neural net]
We say that a neural net \(N\) on a set \(A\) is \emph{discrete} when \(A\) is a finite set.
\end{defn}

Our general definition also encompasses neural nets on infinite sets, such as \(\R\).

\begin{example}
The neural nets that Cybenko considered in his work on universal approximation\cite{cybenko1989} would, in our scheme, have \(r=5\) layers (one input, three hidden, and one output). We would take \(A=\R\) and allow our activation functions to be either constants (the weights), the identity map \(\id_\R\colon\R^1\to\R\) (for carrying a value forward to the next layer unchanged), a sigmoid function
	\[
		\sigma(t)\coloneqq\frac{1}{1+e^{-t}},
	\]
or a dot product. Other characterizations are also possible, depending on which activation functions one allows.
\end{example}

Note that we count layers \(1\) and \(r\) (the input and output layers, respectively) in the total number of layers in the neural net. Thus, a neural net with no hidden layers has \(2\) layers by this definition, while a neural net with one hidden layer has \(3\) layers by this definition, and so on.

Neural nets represent functions by way of generalized composition.

\begin{defn}[Function represented by a neural net]
Given a neural net
    \[
        N_r\coloneqq(V_1,\dots,V_r,E,\Phi,\le)
    \]
on a set \(A\) the \emph{function represented} by \(N_r\) is
    \[
        g_r\colon A^{\abs{V_1}}\to A^{\abs{V_r}}
    \]
where
	\begin{enumerate}
		\item \(g_r=\id_{A^{\abs{V_1}}}\) when \(r=1\) and
		\item when \(r>1\) we set
		    \[
		        (g_r(x))_j\coloneqq\phi_{r,j}((g_{r-1}(x))_{j_1},\dots,(g_{r-1}(x))_{j_{\rho(v_{r,j})}})
		    \]
		    where \(x=(x_1,\dots,x_{\abs{V_1}})\), \(g_{r-1}\) is the function represented by the neural net \(N_{r-1}\coloneqq(V_1,\dots,V_{r-1},E',\Phi',\le')\) obtained by deleting the \(r^{\text{th}}\) layer of \(N_r\), and the in-neighborhood of \(v_j\) in \((V,E)\) is
		        \[
		            \set{v_{r-1,j_1},v_{r-1,j_2},\dots,v_{r-1,j_{\rho(v_{r,j})}}}
		        \]
		    where
		        \[
		            v_{r-1,j_1}<v_{r-1,j_2}<\cdots<v_{r-1,j_{\rho(v_{r,j})}}.
		        \]
	\end{enumerate}
\end{defn}

Our previous examples of neural nets also give us examples of functions represented by neural nets.

\begin{example}
Let \(N\) be the neural net from \autoref{ex:neural_net}. The function \(g\colon\F_5^3\to\F_5^2\) represented by the neural net \(N\) is given by
    \begin{align*}
        g(x_1,x_2,x_3) &\coloneqq (\phi_{31}(\phi_{21}(x_1),\phi_{22}(x_1,x_2,x_3)),\phi_{32}(\phi_{22}(x_1,x_2,x_3),\phi_{23}(x_3),\phi_{24}())) \\
        &= (\phi_{31}(-x_1,x_1-x_2+x_3),\phi_{32}(x_1-x_2+x_3,-x_3,3)) \\
        &= (-x_1(x_1-x_2+x_3)^2,(x_1-x_2+x_3)(-x_3)-3).
    \end{align*}
It can be helpful to view the architecture of the neural net with each node labeled by its corresponding argument variable (for the input nodes in layer \(1\)) or activation function (for the nodes in subsequent layers), as shown in \autoref{fig:labeled_neural_net_architecture}.

    \begin{figure}
        \centering
        \includegraphics[height=5cm]{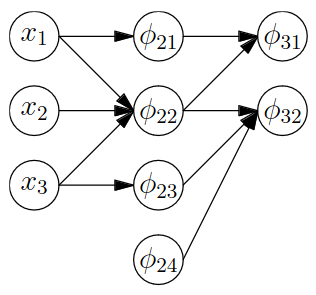}
        \caption{Labeled architecture of a neural net}
        \label{fig:labeled_neural_net_architecture}
    \end{figure}
\end{example}

\section{Learning algorithm}
If discrete neural nets are to have some utility, there ought to be a learning algorithm which tells us how to take a given neural net and modify its activation functions in such a way as to obtain a new neural net which performs better on a given learning task. Traditionally this was done by using differentiable activation functions on a neural net with universe \(\R\) so that a loss function could be differentiated and weights could be adjusted in the direction that would most quickly reduce the empirical loss.

It is immediately apparent that we cannot do something identical, as we have thrown out the notion of continuity, much less differentiability. We can recover some of what we lost by specifying which operations on our universe are ``close to'' each other, as captured in the following definition.

\begin{defn}[Neighbor function, neighbor]
Given a clone \(\mathcal{C}\) we say that a function \(\eta\colon\mathcal{C}\to\pow_\PP(\mathcal{C})\) is a \emph{neighbor function} on \(\mathcal{C}\) when
    \begin{enumerate}
        \item if \(g\in\op_n(\mathcal{C})\) then \(\eta(g)\subset\op_n(\mathcal{C})\) and
        \item for all \(g\in\mathcal{C}\) we have \(g\in\eta(g)\).
    \end{enumerate}
We refer to a member of \(\eta(g)\) as a \emph{neighbor} of \(g\) (with respect to \(\eta\)).
\end{defn}

This definition refers to an arbitrary clone \(\mathcal{C}\) rather than just the clone of all operations \(\op(A)\). The reason for this is that using the clone of all operations will allow us to learn any function given a large enough architecture, but this is exactly what allows overfitting. We will give an example of a sensible family of choices for the clone \(\mathcal{C}\) in the following section.

Given a choice of a neighbor function, we now have an idea of which activation functions are ``close to'' each other but we still don't have an analogue of differentiation which would allow us to pick the best way to improve our neural net's performance by changing an activation function. Here we proceed by simply trying each of the neighbor functions to a given activation function at a particular node, checking whether each one reduces the empirical loss, and then switching the activation function at that node to the one which reduces the empirical loss the most.

A single iteration of our learning algorithm is given in pseudocode in \autoref{alg:learning_algorithm}.

\begin{algorithm}
    \caption{Our learning algorithm}
    \label{alg:learning_algorithm}
    \begin{algorithmic}[1]
    \REQUIRE A neural net \(N\coloneqq(V_1,\dots,V_r,E,\Phi,\le)\) on a set \(A\) whose activation functions all belong to a fixed clone \(\mathcal{C}\).
    \REQUIRE A loss function \(\ell\colon A^{\abs{V_r}}\times A^{\abs{V_r}}\to\R_{\ge0}\).
    \REQUIRE A neighbor function \(\eta\) on \(\mathcal{C}\).
    \REQUIRE A set of training pairs \(T\subset A^{\abs{V_1}}\times A^{\abs{V_r}}\).
    \STATE Select a random node \(v_{ij}\) from \(\bigcup_{s=2}^rV_s\).
    \STATE Create a dictionary \(D\) whose keys are the members of \(\eta(\phi_{ij})\).
    \FORALL{\(g\in\eta(\phi_{ij})\)}
        \STATE Set the value of \(D[g]\) to be \(\frac{1}{\abs{T}}\sum_{(x,y)\in T}\ell(f(x),y)\) where \(f\) is the function represented by the neural net \(N_g\) obtained from \(N\) by redefining \(\Phi(v_{ij})\) to be \(g\).
    \ENDFOR
    \STATE Fix \(g\) to be a member of \(\eta(\phi_{ij})\) so that \(D[g]=\min(\set[D{[g']}]{g'\in\eta(\phi_{ij})})\).
    \ENSURE The neural net \(N_g\) obtained from \(N\) by redefining \(\Phi(v_{ij})\) to be \(g\).
    \end{algorithmic}
\end{algorithm}

Note that because the clone \(\mathcal{C}\) is closed under generalized composition, the function represented by \(N\) and the function represented by \(N_g\) in \autoref{alg:learning_algorithm} will both always belong to \(\mathcal{C}\). 

We give a few examples of neighbor functions. Our first example generally leads to overfitting in practice and is also very computationally expensive as each operation of arity \(n\) has \(\abs{A}^{\abs{A}^n}\) neighbors.

\begin{example}
In that case that \(\mathcal{C}=\op(A)\) we can define a neighbor function \(\eta\) on \(\mathcal{C}\) by setting \(\eta(g)\coloneqq\op_n(A)\) when \(g\in\op_n(A)\).
\end{example}

Our next example is a bit more structured.

\begin{example}
\label{ex:finite_field_clone}
Take \(\mathcal{C}\coloneqq\poly(\Z/p\Z)\) where \(\Z/p\Z\) is the cyclic group of prime order \(p\). Since \(\poly_n(\Z/p\Z)=(\F_p^n)^*\) we have that each member \(g\) of \(\poly_n(\Z/p\Z)\) may be represented as a vector \(\tilde{g}\coloneqq(g_1,g_2,\dots,g_n)\in\F_p^n\) where \(g(x)=\tilde{g}\cdot x\). We define a neighborhood function \(\eta\) on \(\mathcal{C}\) by
    \[
        \eta(g)\coloneqq\set[h\in(\F_p^n)^*]{\tilde{h}=(g_1+u_1,\dots,g_n+u_n)\text{ where }u_i\in\set{-1,0,1}}
    \]
when \(g\in\poly_n(\Z/p\Z)\).
\end{example}

While it is often difficult to parametrize the polymorphism clone of a structure in the manner of the previous example, we can obtain a practical family of neighbor functions by knowing some collection of endomorphisms of the given structure.

\begin{example}
\label{ex:endomorphism_twist}
Let \(\mathbf{A}\) be a relational structure, let \(H\subset\enm(\mathbf{A})\) be a collection of endomorphisms of \(\mathbf{A}\), and let \(G\subset\poly(\mathbf{A})\) be a collection of polymorphisms of \(\mathbf{A}\). We define a neighbor function \(\eta\) on \(\poly(\mathbf{A})\) by setting
    \[
        \eta(g)\coloneqq\set[h_{n+1}\circ g{[h_1,\dots,h_n]}]{h_i\in H}\cup G.
    \]
\end{example}

In order to actually make use of the preceding neighbor function in our algorithm we need to know some collection \(G\subset\poly(\mathbf{A})\) of higher-arity polymorphisms of \(\mathbf{A}\) in order to set the initial activation functions for higher-arity nodes in our neural net. Finding such polymorphisms is not a tractable task in general, but we can do this in some specific cases of interest.

\section{Example application: Polymorphisms for binary images}

We consider a special case of the neighbor function given in \autoref{ex:endomorphism_twist}. Fix \(n\in\N\) and take \(A_n\coloneqq\mat_n(\F_2)\). We refer to \(a\in A_n\) as a \emph{binary image} and we say that \(a\in A_n\) has \emph{size} \(n\). We refer to \((i,j)\in[n]^2\) as a \emph{pixel} in this context and say that \(a\in A_n\) has an entry of \(a_{ij}\) at pixel \((i,j)\).


There is a natural notion of distance between two binary images.

\begin{defn}[Hamming distance]
Given binary images \(a_1,a_2\in A_n\) the \emph{Hamming distance} between \(a_1\) and \(a_2\) is
    \[
        d(a_1,a_2)\coloneqq\left\lvert\set[(i,j)\in{[n]}^2]{(a_1)_{ij}\neq(a_2)_{ij}}\right\rvert.
    \]
\end{defn}

We can use these distances to create a combinatorial graph whose vertices are binary images.

\begin{defn}[Hamming graph]
Given \(n\in\N\) we define the \emph{\(n\)-Hamming graph} to be
    \[
        \ham_n\coloneqq(A_n,\set[(a_1,a_2)\in A_n^2]{d(a_1,a_2)\le1}).
    \]
\end{defn}

That is, given binary images \(a_1,a_2\in A_n\) we say that \(a_1\sim a_2\) in \(\ham_n\) when either \(a_1=a_2\) or \(a_1\) and \(a_2\) have the same values at all pixels except one. Thus, \(a_1\sim a_2\) when the binary images \(a_1\) and \(a_2\) differ in at most one pixel.

Of course there are many possible variations on the definition of a Hamming graph given here. Note that if we had required that \(d(a_1,a_2)=1\) then we would have the \(n^2\)-cube graph instead. It follows that \(\ham_n\) is isomorphic to the graph whose nodes are the vertices of the \(n^2\)-dimensional cube and whose edges are those of the cube along with a single loop at each vertex.


The loops at each vertex will be useful to us in what follows. In order to apply the ideas of \autoref{ex:endomorphism_twist} in the context where the relational structure in question is \(\ham_n\) we must find a set of endomorphisms \(H_n\subset\enm(\ham_n)\) and a set of higher-arity polymorphisms \(G_n\subset\poly(\ham_n)\).

\subsection{Endomorphisms of the Hamming graph}
We give several families of endomorphisms of the Hamming graph \(\ham_n\).

\begin{defn}[Dihedral endomorphism]
Identify \(\mathbf{D}_4\) with the group of isometries of the plane generated by
    \[
        \begin{bmatrix}
            0 & -1 \\
            1 & 0
        \end{bmatrix}
        \text{ and }
        \begin{bmatrix}
            1 & 0 \\
            0 & -1
        \end{bmatrix},
    \]
and define a map \(\gamma_n\colon[n]^2\to U_n\) where
    \[
        U_n\coloneqq\left(\left[-\left\lfloor\frac{n}{2}\right\rfloor,\left\lfloor\frac{n}{2}\right\rfloor\right]\cap\Z\right)^2
    \]
when \(n\) is odd and
    \[
        U_n\coloneqq\set[(x,y)\in U_{n+1}]{x\neq0\text{ and }y\neq0}
    \]
when \(n\) is even by setting
    \[
        \gamma_n(i,j)\coloneqq\left(-\left\lfloor\frac{n}{2}\right\rfloor+j,\left\lfloor\frac{n}{2}\right\rfloor-i\right)
    \]
when \(n\) is odd and setting
    \[
        \gamma_n(i,j)\coloneqq
        \begin{cases}
            \left(-\lfloor\frac{n}{2}\rfloor+j,\lfloor\frac{n}{2}\rfloor-i\right) &\text{when } i,j<\frac{n}{2} \\
            \left(-\lfloor\frac{n}{2}\rfloor+j,\lfloor\frac{n}{2}\rfloor-i-1\right) &\text{when } i\ge\frac{n}{2}\text{ and }j<\frac{n}{2} \\
            \left(-\lfloor\frac{n}{2}\rfloor+j+1,\lfloor\frac{n}{2}\rfloor-i\right) &\text{when } i<\frac{n}{2}\text{ and }j\ge\frac{n}{2} \\
            \left(-\lfloor\frac{n}{2}\rfloor+j+1,\lfloor\frac{n}{2}\rfloor-i-1\right) &\text{when } i,j\ge\frac{n}{2}
        \end{cases}
    \]
when \(n\) is even. Given \(\sigma\in D_4\) define the \emph{dihedral endomorphism} \(h_\sigma\colon A_n\to A_n\) by
    \[
        (h_\sigma(a))_{ij}\coloneqq a_{\gamma_n^{-1}\sigma\gamma_n(i,j)}.
    \]
\end{defn}

Observe that these dihedral endomorphisms are actually automorphisms. We write \(\dihedralend_n\) to indicate the set of dihedral endomorphisms \(\ham_n\). The class of dihedral automorphisms belongs to the larger class of automorphisms obtained by permuting the pixels of an image according to an arbitrary permutation of \([n]^2\), but we will restrict ourselves to \(\dihedralend_n\) as these automorphisms are easier to store and compute with than a general permutation.

Another class of automorphisms are given by adding a fixed binary image pointwise.

\begin{defn}[Swapping endomorphism]
Given \(b\in A_n\) the \emph{swapping endomorphism} for \(b\) is the map \(h_b^+\colon A_n\to A_n\) which is given by \(h_b^+(a)\coloneqq a+b\) where the sum \(a+b\) is the usual componentwise sum of matrices over \(\F_2\).
\end{defn}

Note that swapping endomorphisms are also automorphisms. We write \(\swapend_n\) to indicate the set of swapping endomorphisms of \(\ham_n\).

Finally, we have endomorphisms which are not automorphisms. These are obtained by taking the Hadamard product with a fixed binary image.

\begin{defn}[Blanking endomorphisms]
Given \(b\in A_n\) the \emph{blanking endomorphism} for \(b\) is the map \(h_b^\odot\colon A_n\to A_n\) which is given by \(h_b^\odot(a)\coloneqq a\odot b\) where \(a\odot b\) is the Hadamard product of matrices over \(\F_2\).
\end{defn}

We denote by \(\blankend_n\) the set of all blanking endomorphisms of \(\ham_n\).

By setting
    \[
        H_n\coloneqq\dihedralend_n\cup\swapend_n\cup\blankend_n
    \]
we take our given set of endomorphisms to consist of all those from the three above-defined classes.

\subsection{Polymorphisms of the Hamming graph}
In this subsection we will make use of a notion related to that of Hamming distance.

\begin{defn}[Hamming weight]
Given a binary image \(a\in A_n\) the \emph{Hamming weight} of \(a\) is
    \[
        \norm{a}\coloneqq d(a,0)
    \]
where \(0\) is the image whose pixels are all \(0\).
\end{defn}

Thus, \(\norm{a}\) is the number times \(1\) appears as an entry of \(a\). The following set will be useful.

\begin{defn}[Standard basis]
The \emph{standard basis} of the space of size \(n\) binary images is
    \[
        B_n\coloneqq\set[a\in A_n]{\norm{a}=1}.
    \]
\end{defn}

This is to say that \(B_n\) consists of all images where exactly one pixel is given the value \(1\). Note that \(a_1\sim a_2\) in \(\ham_n\) if and only if \(a_1+a_2\in B_n\cup\set{0}\).

Any function \(h\colon A_n^k\to A_n\) with \(\im(h)=\set{a_1,a_2}\) with \(a_1\sim a_2\) is a polymorphism. These are unwieldy to work with and store in general so we restrict our attention to a special class of such polymorphisms.

\begin{defn}[Multi-linear indicator]
Given \(b\in B_n\) and \(c\in A_n^k\) the \emph{multi-linear indicator polymorphism} for \((b,c)\) is the map \(g_{b,c}\colon A_n^k\to A_n\) given by
    \[
        g_{b,c}(a_1,\dots,a_k)\coloneqq\left(\prod_{i=1}^ka_i\cdot c_i\right)b
    \]
where \(x\cdot y\coloneqq\sum_{i,j}x_{ij}y_{ij}\) denotes the standard dot product in \(\F_2^{[n]^2}\).
\end{defn}

We denote the class of multi-linear indicator polymorphisms of \(\ham_n\) by \(\multind_n\).

Our final, most interesting, class of polymorphisms has members whose images are not generally a pair of adjacent vertices in the Hamming graph. In order to define these we need to set up some machinery.

\begin{defn}[Hamming weight map]
The \emph{\(k\)-ary Hamming weight map}
    \[
        \psi_k\colon A_n^k\to[n^2+1]^k
    \]
is given by
    \[
        \psi_k(a_1,\dots,a_k)\coloneqq(\norm{a_1},\dots,\norm{a_k}).
    \]
\end{defn}

This map induces a relation on the members of \(A_n^k\), which is \(\ker(\psi_k)\subset(A_n^k)^2\). We consider the quotient of \(\ham_n^k\) induced by this relation.

\begin{defn}[Hamming weight graph]
We refer to the graph \(\ham_n^k/\ker(\psi_k)\) as the \emph{\(k\)-ary Hamming weight graph of size \(n\)}.
\end{defn}

Observe that \(\ham_n^k/\ker(\psi_k)\) is isomorphic to the graph whose vertex set is \([n^2+1]^k\) and whose edges are between \(k\)-tuples \((u_1,\dots,u_k)\) and \((v_1,\dots,v_k)\) such that for all \(i\) we have \(\abs{u_i-v_i}\le1\). We will suppress this isomorphism in what follows. We denote the canonical homomorphism of graphs from \(\ham_n^k\) to the quotient \(\ham_n^k/\ker(\psi)\) by \(\tilde{\psi}\).

Our strategy for finding polymorphisms of \(\ham_n\) is as follows. We will give a procedure for finding graph homomorphisms \(f\colon\ham_n^k/\ker(\psi_k)\to\ham_n\). Given such a homomorphism we will obtain a polymorphism \(f\circ\tilde{\psi}\colon\ham_n^k\to\ham_n\) by composing with the quotient map \(\tilde{\psi}\). Since the Hamming weight graph is simpler to understand than powers of this original Hamming graph this will simplify our search.

In order to define such homomorphisms \(f\colon\ham_n^k/\ker(\psi_k)\to\ham_n\) we introduce the following combinatorial objects.

\begin{defn}{Basic cube}
A \emph{basic cube} \(C_u\) of the \(k\)-ary Hamming weight graph of size \(n\) is a set of vertices
    \[
        \set[(v_1,\dots,v_k)\in{[n^2+1]}^k]{(\forall i)(v_i-u_i\in\set{0,1})}
    \]
where \(u=(u_1,\dots,u_k)\in[n^2]^k\) is called the \emph{top corner} of the basic cube.
\end{defn}

We will denote by \(\mathbf{K}_L\) the complete graph on the vertex set \(L\) with a loop at each vertex. Our (slightly nonstandard) definition of an \emph{\(L\)-coloring} of a graph \(\mathbf{G}\) is then a homomorphism from \(\mathbf{G}\) to \(\mathbf{K}_L\). Note that under this weakened definition any function from the vertex set of \(\mathbf{G}\) to \(L\) would constitute an \(L\)-coloring of \(\mathbf{G}\).

\begin{defn}[Dominion]
Given a set of labels \(L\) a \emph{\((k,n,L)\)-dominion} is an \(L\)-coloring of the \(k\)-ary Hamming weight graph of size \(n\) such that given any vertex \(u\in[n^2]^k\) the basic cube \(C_u\) is colored using at most \(2\) members of \(L\).
\end{defn}

We will often refer to a \((k,n,L)\)-dominion as the corresponding function
    \[
        D\colon\ham_n^k/\ker(\psi_k)\to L.
    \]

There is a graph whose vertices are the labels \(L\) induced by such a coloring.

\begin{defn}[Minimum constraint graph]
Given a \((k,n,L)\)-dominion \(D\) the \emph{minimum constraint graph} \(\minc(D)\) is the simple graph whose vertices are \(L\) and whose adjacency relation is given by \(\ell_1\sim \ell_2\) when there is a basic cube \(C_u\) of \(\ham_n/\ker(\psi_k)\) such that the labels \(\ell_1\) and \(\ell_2\) are both used to color at least one vertex from \(C_u\).
\end{defn}

Note that the minimum constraint graph of a \((k,n,L)\)-dominion \(D\) is the subgraph of \(\mathbf{K}_L\) whose edges consist only of those joining pairs of distinct vertices in \(\im(D)\).

Homomorphisms from the minimum constraint graph of a dominion to \(\ham_n\) yield polymorphisms of \(\ham_n\).

\begin{defn}[Dominion polymorphism]
Given a \((k,n,L)\)-dominion \(D\) and a homomorphism \(\alpha\colon\minc(D)\to\ham_n\) the \emph{dominion polymorphism} \(g_\alpha\colon\ham_n^k\to\ham_n\) is given by
    \[
        g_\alpha(a_1,\dots,a_k)\coloneqq\alpha(D(\psi_k(a))).
    \]
\end{defn}

Given a typical dominion \(D\) it may be challenging to find homomorphisms \(\alpha\colon\minc(D)\to\ham_n\). If we assume that our minimum constraint graph is a subgraph of a tree we can find such homomorphisms more easily.

Given a \((k,n,L)\)-dominion \(D\) and any graph \(\mathbf{L}\) which contains \(\minc(D)\) as a subgraph, we know that there is an inclusion homomorphism \(\iota\colon\minc(D)\hookrightarrow\mathbf{L}\). It follows that if we can find a homomorphism \(\alpha'\colon\mathbf{L}\to\ham_n\) then we can take
    \[
        \alpha\coloneqq\alpha'\circ\iota\colon\minc(D)\to\ham_n
    \]
as our homomorphism for creating a dominion polymorphism
    \[
        g_\alpha\colon\ham_n^k\to\ham_n.
    \]

Code for this example can be found on GitHub\footnote{\href{https://github.com/caten2/Tripods2021UA}{https://github.com/caten2/Tripods2021UA}}. At the time of this writing this repo is still under active development.

We found that neural nets equipped with polymorphisms of the Hamming graph as activation functions would converge quickly to a minimum possible empirical loss. It is important to note that this final empirical loss was not \(0\), as such overfitting is made impossible by our choice of activation function.

\printbibliography

\end{document}